\documentclass[conference]{IEEEtran}
\IEEEoverridecommandlockouts
\usepackage{cite}
\usepackage{amsmath,amssymb,amsfonts}
\usepackage{algorithmic, subfigure}
\usepackage{graphicx}
\usepackage{textcomp}
\usepackage{xcolor}
\usepackage{mathptmx} 
\usepackage{tabularx,multirow,makecell,url,float,bm,color,subcaption}
\usepackage{marvosym,threeparttable,booktabs}
\usepackage[switch]{lineno}
\usepackage[ruled,vlined]{algorithm2e}
\usepackage[pagebackref,breaklinks,colorlinks,
          linkcolor = black,	
          anchorcolor = black,
          urlcolor = black,	
          citecolor = black,bookmarks=false]{hyperref}

\newcommand{\figref}[1]{\mbox{Figure~\ref{#1}}}
\newcommand{\tabref}[1]{\mbox{Table~\ref{#1}}}

\def\BibTeX{{\rm B\kern-.05em{\sc i\kern-.025em b}\kern-.08em
    T\kern-.1667em\lower.7ex\hbox{E}\kern-.125emX}}
\begin{document}

\title{EPIC: Efficient Prompt Interaction for Text-Image Classification}

\author{
    \IEEEauthorblockN{
        Xinyao Yu$^1$\IEEEauthorrefmark{1}, 
        Hao Sun$^1$\IEEEauthorrefmark{1}, 
        Zeyu Ling$^1$, 
        Ziwei Niu$^1$, 
        Zhenjia Bai$^{1}$, 
        Rui Qin$^1$,
        Yen-Wei Chen$^2$\IEEEauthorrefmark{2},
        Lanfen Lin$^1$\IEEEauthorrefmark{2}
    }
    \vspace{0.1cm}
    
    \IEEEauthorblockA{
        $^1$College of Computer Science and Technology, Zhejiang University, Hangzhou, China
    }
    \IEEEauthorblockA{
        $^2$College of Information Science, Ritsumeikan University, Shiga, Japan\\
    \vspace{0.1cm}
    Email: xinyaoyu@zju.edu.cn, llf@zju.edu.cn}
    \thanks{$^*$Equal contribution. $^{\dag}$Corresponding authors.}
}

\maketitle

\begin{abstract}
In recent years, large-scale pre-trained multimodal models (LMMs) generally emerge to integrate the vision and language modalities, achieving considerable success in multimodal tasks, such as text-image classification. The growing size of LMMs, however, results in a significant computational cost for fine-tuning these models for downstream tasks. Hence, prompt-based interaction strategy is studied to align modalities more efficiently. In this context, we propose a novel efficient prompt-based multimodal interaction strategy, namely \textbf{E}fficient \textbf{P}rompt \textbf{I}nteraction for text-image \textbf{C}lassification (\textbf{EPIC}). Specifically, we utilize temporal prompts on intermediate layers, and integrate different modalities with similarity-based prompt interaction, to leverage sufficient information exchange between modalities. Utilizing this approach, our method achieves reduced computational resource consumption and fewer trainable parameters (about 1\% of the foundation model) compared to other fine-tuning strategies. Furthermore, it demonstrates superior performance on the UPMC-Food101 and SNLI-VE datasets, while achieving comparable performance on the MM-IMDB dataset. 
\end{abstract}


\section{Introduction}
\label{sec:intro}

Modern Internet platforms, encompassing social media and e-commerce, host a myriad of content expressed across various modalities, predominantly vision and language. Harnessing information from different modalities has demonstrated its potential to enhance performance on diverse multimodal tasks, such as text-image classification~\cite{fu2022cma}, recommendation~\cite{lu2021future}, and sentiment analysis~\cite{deng2021dense}. Multimodal learning commonly employs a strategy of interacting representations from involved modalities. Previous multimodal interaction methods~\cite{zadeh2017tensor,zadeh2018memory,hazarika2020misa,vielzeuf2018centralnet} involves extracting unimodal features separately with distinct backbones, and then blends these unimodal representations using a fusion module. However, this approach faces challenges in fully addressing comprehensive inter- and intra-modality relationships, resulting in insufficient inter-modality interaction.

\begin{figure*}
\centering
\begin{minipage}{0.32\linewidth}
\centering
\includegraphics[width=1\textwidth]{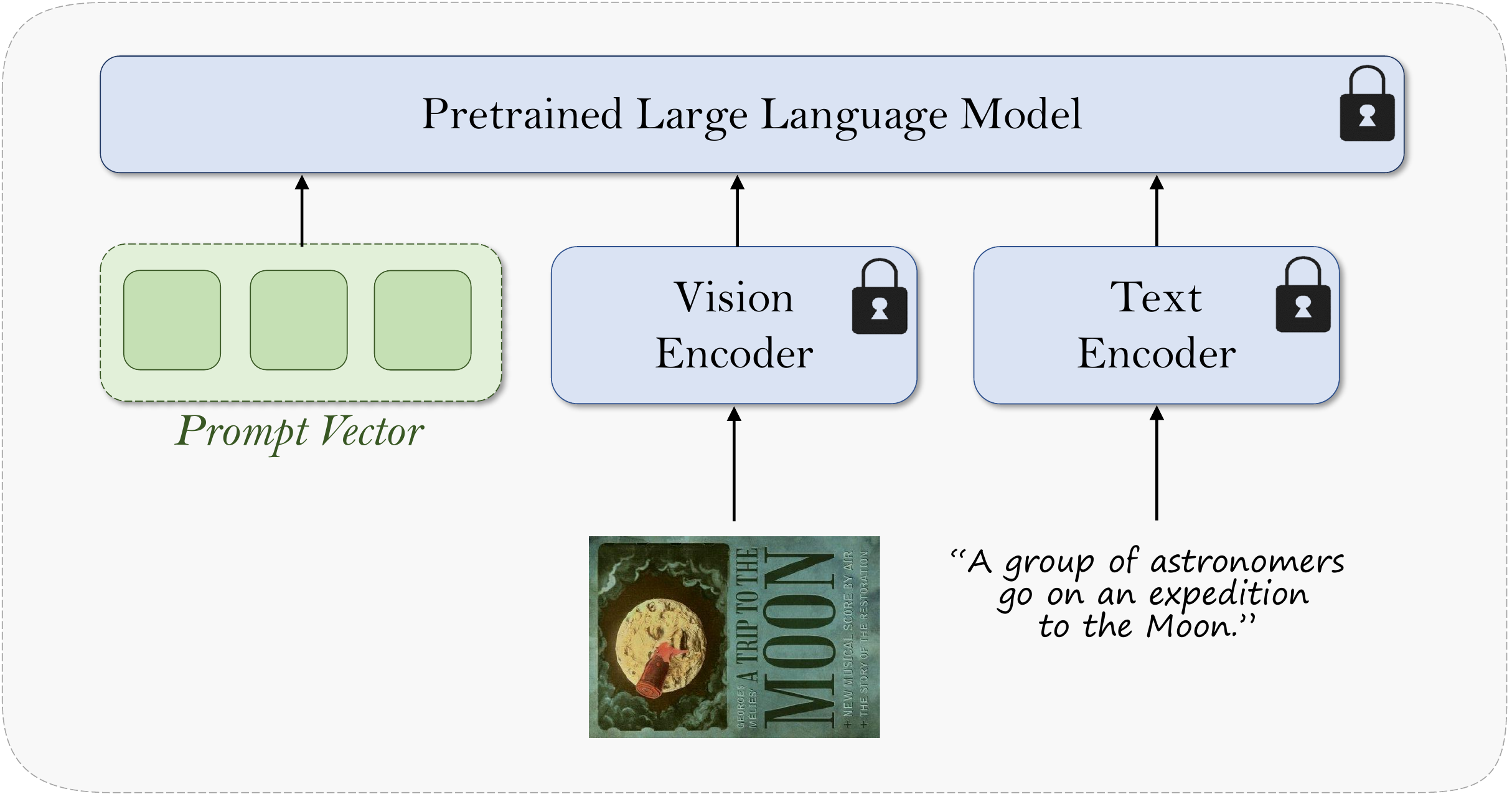}
\caption*{(a) The first proposed prompt-based interaction strategy.\cite{liang2022modular}}
\label{fig1a}
\end{minipage}
\begin{minipage}{0.32\linewidth}
\centering
\includegraphics[width=1\textwidth]{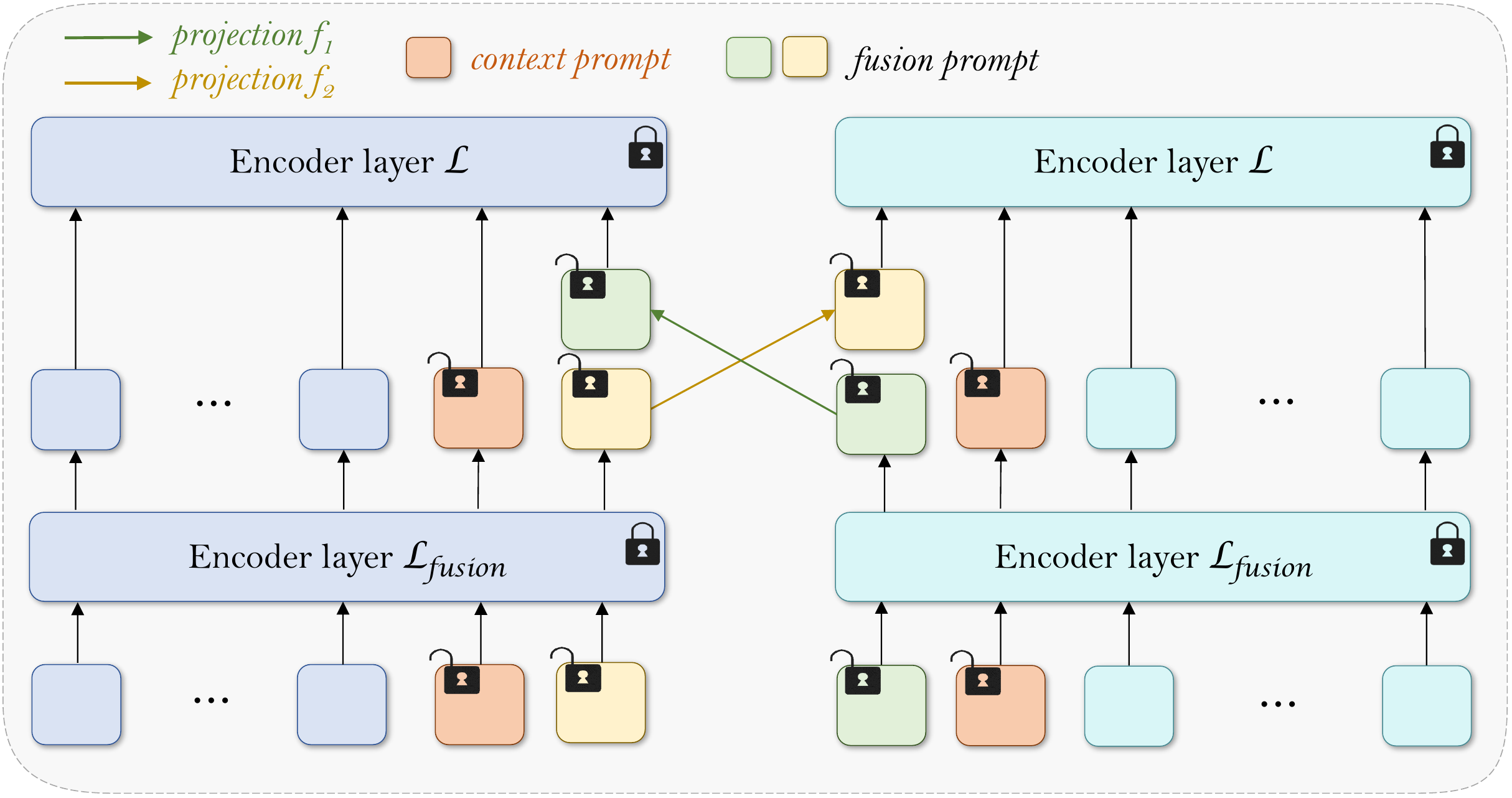}
\caption*{(b) Existing prompt-based interaction strategy.\cite{li2023efficient}}
\label{fig5b}
\end{minipage}
\begin{minipage}{0.32\linewidth}
\centering
\includegraphics[width=1\textwidth]{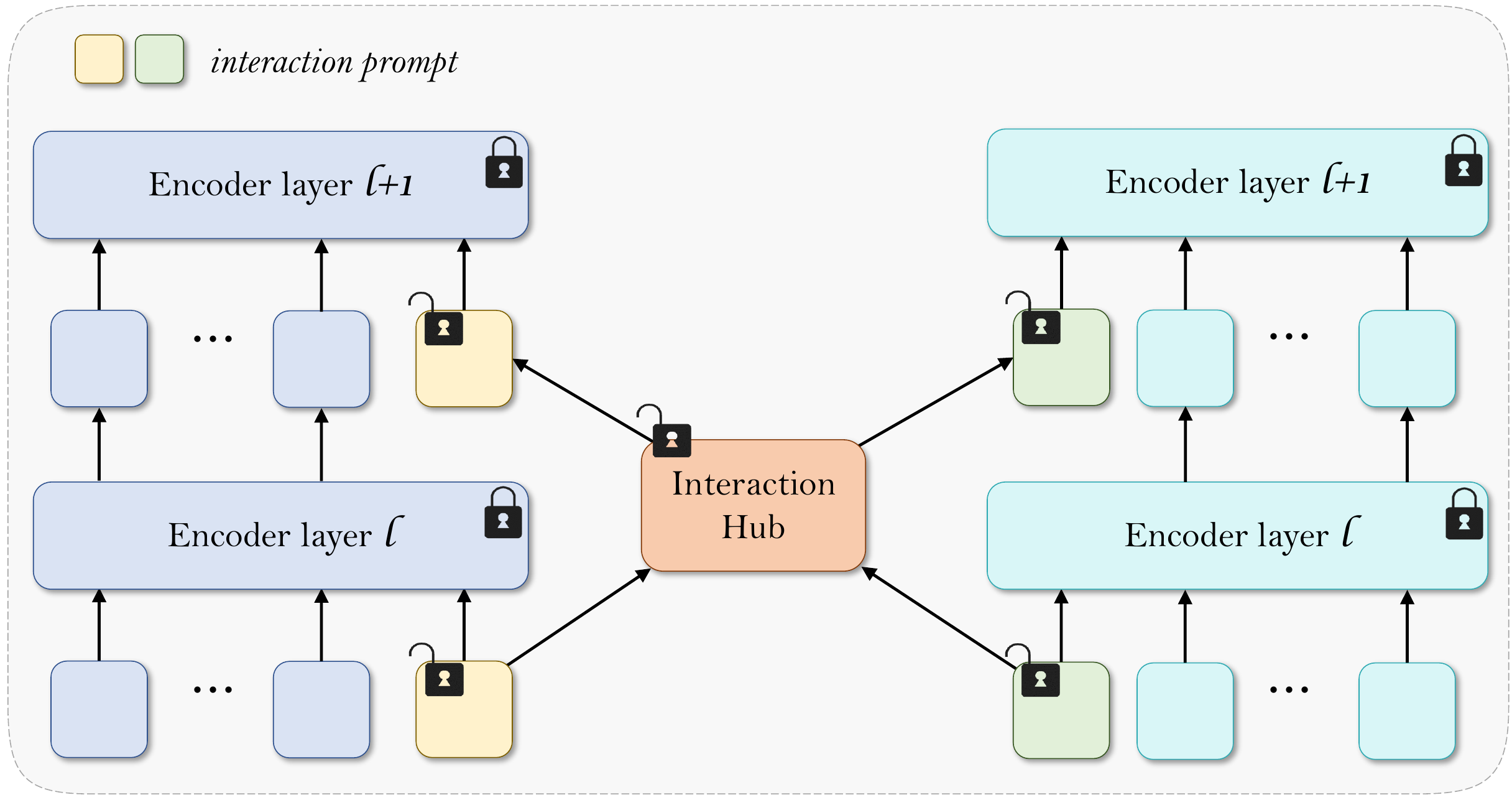}
\caption*{(c) Our proposed prompt interaction strategy EPIC.}
\label{fig5c}
\end{minipage}
\caption{Comparisons among existing prompt-based interaction strategies (a)/(b), and our proposed strategy (c). (a) extracts unimodal representations separately and then utilizes prompt vectors to align the extracted representations with PLM. (b) leverages prompt vectors for each branch of the model, and utilizes the prompts to deliver information from the other branch on intermediate layers. In our proposed EPIC (c), prompt vectors are leveraged for each modality on intermediate layers, and are then interacted to produce prompts of the next layer. Direct interactions of prompts from different branches allow interactions between modalities.} 
\label{fig1}
\end{figure*}
 
As large-scale multimodal models (LMMs) gain prominence in multimodal learning, fine-tuning methods~\cite{kiela2019supervised,jia2022visual,zhou2022learning,zhou2022conditional,khattak2023maple} for downstream tasks become prevalent, albeit with a significant drawback of large memory usage. To mitigate this issue when exploiting pre-trained models in downstream tasks, a prompt-based cross-modal interaction strategy~\cite{liang2022modular,li2023efficient} has emerged, albeit still in its infancy, as depicted in~\figref{fig1} (a) and (b). The prompt-based strategy leverages prompt vectors to either align different modalities in prefix manner, or facilitate modality interaction on intermediate layers and mid-level features of each modality. 
The former approach allows information exchange with the prefix prompt vectors, however it leads to insufficient information exchange due to the prefix manner alignment of modalities. The latter approach enables inter-modality information exchange through trainable prompts, reducing memory usage due to the low spatial complexity of prompt operations. However, the latter approach lacks direct interaction between trainable prompts, leading to insufficient interaction between modalities. Moreover, the latter approach disentangles prompt vectors into several types, resulting in complexity in method. 

To handle these problems, we propose a novel multimodal interaction method, namely \textbf{E}fficient \textbf{P}rompt \textbf{I}nteraction for text-image \textbf{C}lassification (EPIC). The proposed prompt-based interaction strategy is as shown in~\figref{fig1} (c). The proposed method utilizes temporal prompts on intermediate layers to acquire context information, and then temporal prompts are fed into an Interaction Hub for prompt interaction. In Interaction Hub, we calculate similarity between temporal prompts of each modality to highlight prompt logits of importance as the activation vector of prompts from each modality, and add information from the activated prompt of the other modality while retaining information from prompt of current modality. The interaction of temporal prompts enables a two-way information flow between both modalities without participation of modality features. Note that only the parameters of the prompt tokens and projection block in the Interaction Hub are trainable, which significantly reduces the quantity of trainable parameters and memory usage during training. Despite the variety of downstream tasks in multimodal learning, we focus on text-image classification in this paper. 

In summary, our contribution can be shown as follows:
\begin{enumerate}
    \item We introduce EPIC, a novel prompt-based LMM fine-tuning strategy for multimodal classification tasks, which ensures ample cross-modal information flow through a lightweight, similarity-based multimodal information Interaction Hub.  
    \item We propose the Interaction Hub, a similarity-based lightweight module for cross-modal information interaction that effectively compresses the required quantity of training parameters for cross-modal information exchange. 
    \item Extensive experimentation demonstrates that our method achieves state-of-the-art (SOTA) performance among prompt-based strategies on the UPMC-Food101 and SNLI-VE datasets, and attains a level comparable to SOTA among prompt-based strategies on the MM-IMDB dataset. Moreover, we conduct a comparative analysis of computational resource consumption with other fine-tuning schemes. The results indicate that our method requires fewer training parameters compared to other strategies, and exhibits minimal memory cost during training.
\end{enumerate}

	\section{Our Method}\label{sec:method}
	\begin{figure*}
    \centering
    \includegraphics[width=1\linewidth]
    {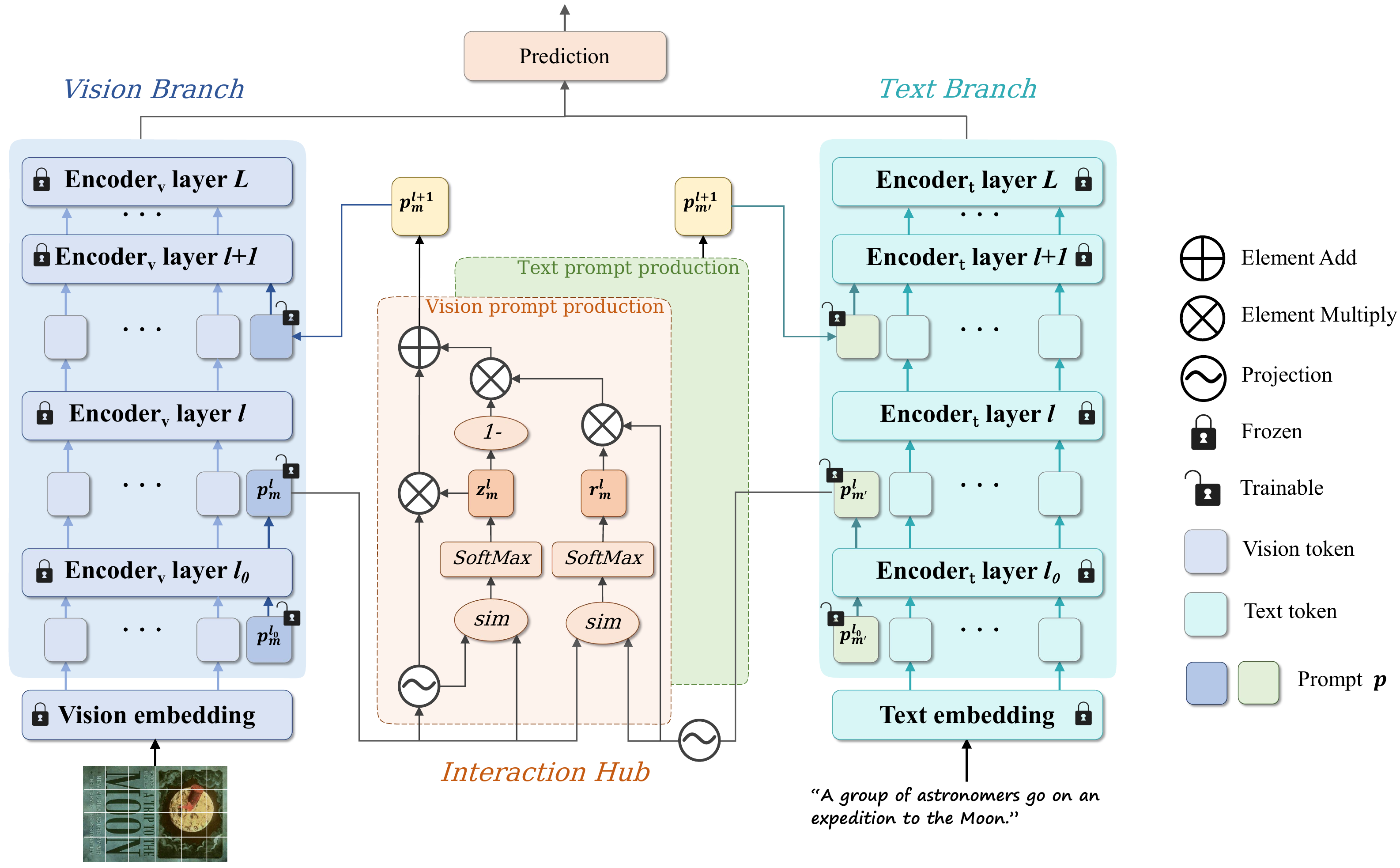}
    \caption{The pipeline of our proposed method. We utilize pre-trained foundation model in frozen for basic feature extraction of image and text branches, and leverage temporal prompts on intermediate layers to store information on temporal layer and act as media for information exchange. Temporal prompts of different modalities are interacted in Interaction Hub to produce prompts for the next layer. 
    $\{m, m'\} \in \{v,t\}$ denotes the vision and text modality, where $m \neq m'$. }
    \label{fig2}
\end{figure*}



\subsection{Problem Statement}

The definition of the image-text pair classification task is as follows: given an image-text pair \(\mathit{X_{\mathit{i} } }=\left \{ \mathcal{V_{\mathit{i} }}, \mathcal{T_{\mathit{i} } }\right\}\) containing an image \(\mathcal{V_{\mathit{i} }}\in \mathbb{R^{\mathit{c\times h\times w} } } \) and a textual description \(\mathcal{T_{\mathit{i} }} \in \mathbb{R}^{\mathit{d_{i}}}\), where $c$ denotes channel, $h$ and $w$ represent the height and width of the image, respectively,, and $d_{i}$ denotes the length of text. \(\mathit{Y_{i}} \in \{1,2,...,K\} \) denotes the class label. The goal is to learn a classifier to predict the class label \(\mathit Y_{i} \) of each image-text pair $\left\{\mathcal{V_{\mathit{i} }}, \mathcal{T_{\mathit{i} } }\right\}$.

\subsection{Model Overview}

The pipeline of the model is as shown in~\figref{fig2}. This work adopts LMM in frozen as foundation model to extract base features of each modality in each branch. Temporal prompts are leveraged on several intermediate layers for intra- and inter-modality information exchange. Temporal prompts serve simultaneously in two roles: first, they engage in self-attention operations with tokens of the same modality, storing contextual information of the current modality; second, they participate in computations within the Interaction Hub, facilitating cross-modal information exchange with other modal branches. 
In practice, we select several layers to perform prompt interaction, denoted as interaction layer, while those layers without prompts are denoted as extraction layer.

\subsection{Temporal Prompt}
The adopted LMM foundation model contains a visual branch \(Encoder_{v}\) and a textual branch \(Encoder_{t}\). Given a pair of image and tokenized text \(\left \{ \mathcal{V_{\mathit{i} },T_{\mathit{i}} } \right\} \), we first feed the image \(\mathcal{V_{\mathit{i} }} \) into \(Encoder_{v}\) and the tokenized text \(\mathcal{T_{\mathit{i} }} \) into \(Encoder_{t}\) for basic feature extraction. The image feature and text feature on intermediate layer \(l\) are denoted as \(u_{m}^{l}\), where \(\mathit m \in \left\{v,t \right\} \) denotes modality. Temporal prompts \(\mathit{p_{m}^{l} } \) are introduced on interaction layers of each modality branch, where \(\mathit m \in \left\{v,t \right\} \) denotes modality and \(\mathit l \) refers to the interaction layer.  

We designate the first interaction layer as layer \(\mathit l_{0}\). Following previous work~\cite{khattak2023maple}, we first randomly initialize the text prompt \(\mathit{p_{t}^{l_{0}} } \) with Gaussian distribution, and utilize it to produce vision prompt. The first temporal prompt for text is initialized to be a trainable vector. The first temporal prompt for image branch \(\mathit{p_{v}^{l_{0}} } \) is obtained via linear transformation from the temporal prompt of the text branch through MLP projection, as follows:

\begin{equation}
    p_{v}^{l_{0}} = W \cdot p_{t}^{l_{0}} + b \label{XX1},
\end{equation}
where \(W\) and \(b\) respectively denote the learnable weight and bias of the linear layer.
On each interaction layer of branch $Encoder_{m}$, we first concatenate the modality feature input \({u}_{m}^{l}  \) and the corresponding temporal prompt \({p_{m}^{l} } \). $p_{m}^{l}$ provides information from previous layers of both modalities, and acts as temporal prompt for layer $l$ of branch $m$. Then the concatenated input is fed into the pre-trained encoder layer \(l\) as:
\begin{equation}
    [\hat{p}_{m}^{l}; {u}_{m}^{l+1} ]=Encoder_{m}^{l}([p_{m}^{l}; {u}_{m}^{l}]), \label{XX2}
\end{equation}
where \([\cdot; \cdot] \) refers to concatenation operation, and $[\hat{p}_{m}^{l}; {u}_{m}^{l+1} ]$ is the output, and $\hat{p}_{m}^{l}$ will be leveraged for inter-modality interaction in the interaction stage. Through this process, context information of layer $l$ from modality $m$ is acquired in temporal prompt. Similarly, the input for the next layer is $[p_{m}^{l+1}; {u}_{m}^{l+1}]$. Each interaction layer is associated to an Interaction Hub, to integrate the information from both modalities and produce temporal prompts ${p_{v}^{l+1},p_{t}^{l+1}}$ for the next layer, as:
\begin{equation}
    \left \{p_{v}^{l+1},p_{t}^{l+1} \right \} = \text{InteractionHub}(\hat{p}_{v}^{l},\hat{p}_{t}^{l}). \label{xx3}
\end{equation}


\subsection{Interaction Hub} \label{memory_hub}

In this section, we introduce the Interaction Hub, which implements the prompt interaction strategy.

We first project $\hat{p}_{m'}^{l}$ into the space of $\hat{p}_{m}^{l}$ through a two-layer MLP, denoted as $\tilde p_{m'}^{l}$. We consider two types of correspondences between temporal prompts: those intra-modality and those inter-modality. Analogical mapping connection is leveraged to discover correspondences between tokens, working as $Mapping(A,B)=f_{M}(sim(A,B))$, where $f_{M}$ means the mapping function, and $sim$ means similarity function. Specifically, the correspondence mapping connections are calculated as:
\begin{equation}
    \begin{aligned}
        F_{m}^{intra} & = ReLU(sim(MLP(\hat{p}_{m}^{l}),\hat{p}_{m}^{l})), \\
        F_{m}^{inter} & = ReLU(sim(\hat{p}_{m}^{l}, \tilde p_{m'}^{l})).
        \label{XX4}
    \end{aligned}
\end{equation}

\begin{table*}[]
\caption{Results on employed datasets compared with other methods. * means that we re-implement the method under our settings, with CLIP backbone. 
}
\label{tab:my_label1}
\centering
\renewcommand\arraystretch{1.0}
\begin{tabular}{lcccccc}
\hline
\multirow{2}{*}{Methods}                & Trainable     & Training/Inference            &  UPMC-Food101   &  MM-IMDB        &  SNLI-VE   &  \multirow{2}{*}{Avg.}\\
                                        & Param.(Million) &  Mem.                 &  Acc(\%)        &  F1-micro/macro &  Acc($\%$) \\ \hline
 HUSE\cite{narayana2019huse}            & - & -  &  92.30 & -      & -         & -\\
 VisualBERT\cite{jia2022visual}         & - & -        & 92.30          & -         &   \textbf{75.06}         &\\
 Late Fusion\cite{liang2021multibench}  & - & -   & -          &  59.6 / 51.0             & -         & -\\
 DynMM\cite{xue2023dynamic}             & - & -  & -          &  61.0 / 51.6             & -         & -\\
 UniT\cite{hu2021unit}                  & - & -  & -          & -              & 73.16         & -\\
 CMA-CLIP\cite{fu2022cma}               & - & -  & 93.10      &  65.3 / 52.7              & -         & -\\ 
 VilT\cite{pmlr-v139-kim21k}            & - & -  &  92.90      & -              &  -         & -\\
 MMBT\cite{kiela2019supervised}         & 196.5 & 37.87 / 3.48  &  92.10      &  \textbf{66.8} / \textbf{61.8}    &  74.69        & \textbf{77.03}\\ \hline
 MaPLe*\cite{khattak2023maple}          & 0.8   & 47.40 / 6.22   & 90.80      & 60.9 / 51.2    & 71.52     & 72.79\\
 P-CLIP                                 & 0.1   & 47.20 / 6.20  & 90.30      & 60.0 / 50.6    & 70.56  & 72.05         \\ 
 PromptFuse\cite{liang2022modular}      & 0.1   & 29.57 / 3.55  & 82.21      & 54.5 / 48.6    & 64.53        & 66.09 \\
 BlindPrompt\cite{liang2022modular}     & 0.1   & 29.57 / 3.65 & 84.56      & 56.5 / 50.2    & 65.54    &  67.81\\
 PMF*\cite{li2023efficient}              & 2.7   & 15.29 / 6.44  & 91.96      & 64.0 / 58.4   & 72.14         & 75.10\\ 
 PMF-large*\cite{li2023efficient}        & 5.0   & 20.07 / 7.01  & 92.54     & 65.4 / \textbf{60.2}    & 72.40         & 75.91 \\ \hline
 EPIC-base(ours)                        & 2.0   & 17.63 / 6.28 & \textbf{93.95} & \textbf{65.9} / 56.3         & \textbf{73.45}    & \textbf{76.17}    \\
 EPIC-small(ours)                       & 2.0   & \textbf{11.35} / 6.25 &  93.17          & 64.8 / 55.4              & 72.51     & 75.26   \\ \hline
\end{tabular}
\end{table*}

 We leverage $F_{m}^{intra}$ to attain intra-modality activation vector $z_{m}$, which highlights important logits of $\hat{p}_{m}^{l}$, and decides the extent to retain information in $\hat{p}_m^l$, so as to consolidate the context information in $\hat{p}_{m}^{l}$. Also, we leverage $F_{m}^{inter}$ to attain inter-modality activation vector $r_{m}$, which highlights logits of $\tilde p_{m'}^{l}$ with high correspondence to $\hat{p}_{m}^{l}$. In our work, activation of each logit is considered on the basis of mapping connections $F_{m}^{intra}$ and $F_{m}^{inter}$, because the similarity-based mapping connections represent correspondence relationships. Moreover, we consider feature unit activation in soft max manner. Specifically, the intra- and inter-modality activation gate of temporal prompts $z_{m}^l$ and $r_{m}^l$ for modality $m$ are calculated as:
\begin{equation}
    \begin{aligned}
        z_{m}^l &= SoftMax(F_{m}^{intra}),\\
        r_{m}^l &= SoftMax(F_{m}^{inter}),
        \label{XX5}
    \end{aligned}
\end{equation}
where $z_{m}^l$ denotes the intra-modality activation gate of $\hat{p}_{m}^{l}$, and $r_{m}^l$ denotes the inter-modality activation gate of $\tilde p_{m'}^{l}$. Finally we integrate $\hat{p}_{m}^{l}$ and $\tilde p_{m'}^{l}$ to produce $p_{m}^{l+1}$, as:
\begin{equation}
    p_{m}^{l+1} = z_{m}^l \cdot \hat{p}_{m}^{l}+(1-z_{m}^l) \cdot r_{m}^l \cdot \tilde {p}_{m'}^{l}.
    \label{XX6}
\end{equation}
$p_{m'}^{l+1}$ is produced by the same process as above. Through the integration process, $z_m^l$ decides the extent to retain context information in $\hat{p}_m^l$, $(1-z_m^l)$ decides the extent to integrate information from the other modality, and $r_m^l$ reflects the correspondence between $\hat{p}_{m'}^l$ and $\hat{p}_m^l$, deciding the activation rate of $\hat{p}_{m'}^l$. From this point of view, EPIC is able to leverage information from both modalities without disentangling prompt vectors. Note that the trainable parameters are shared in all interaction layers, so that the number of trainable parameters remains the same when interaction layers pile up.

\subsection{Training}

For classification, we add hand-crafted text prompts with class labels \(y \in \left\{1,2,...,K \right\} \) having \(K\) classes. The prediction is made by the similarity score with temperature \(\tau\) between the final output of the vision encoder \(x\) and the text encoder \(h\). The prediction logits \(\hat{y}=(\hat{y}_{1},...,\hat{y}_{K} )\) is calculated as:
\begin{equation}
    p(\hat{y} \mid x)=\frac{\mathrm{exp}(sim(x,h_{\hat{y}})/\tau)  }{\sum_{1}^{K} \mathrm{exp}(sim(x,h_{i})) }, 
    \label{XX7}
\end{equation}
where $i$ refers to each logit.


As we conduct experiments on uni-label classification task and multi-label classification task, we define loss for both tasks.
For uni-label classification task, we define the loss as:
\begin{equation}
    L_{uni} = \frac{1}{N}\sum_i- [y_i\dot log(p_i)+(1-y_i)\dot log(1-p_i)],  \label{XX8}
\end{equation}
where \(y_{i} \in \left\{0,1 \right\}\) is the ground truth label, $p_i$ denotes the predicted probability of sample $i$ to be true.
For multi-label classification task, we define the loss as:
\begin{equation}
    L_{multi} = -\frac{1}{N}\sum_i \sum_{c=1}^K y_{ic} log(p_{ic}),    \label{XX9}
\end{equation}
where $N$ refers to the number of samples, $K$ denotes the number of classes, $y_{ic} \in \{0,1\}$ is designated to be the ground truth label of whether sample $i$ belongs to class $c$, and $p_{ic}$ refers to the predicted probability that sample $i$ belongs to class $c$.

 \begin{figure*}
    \centering
    \begin{minipage}{0.27\linewidth}
    \centering
    \includegraphics[width=1\textwidth]{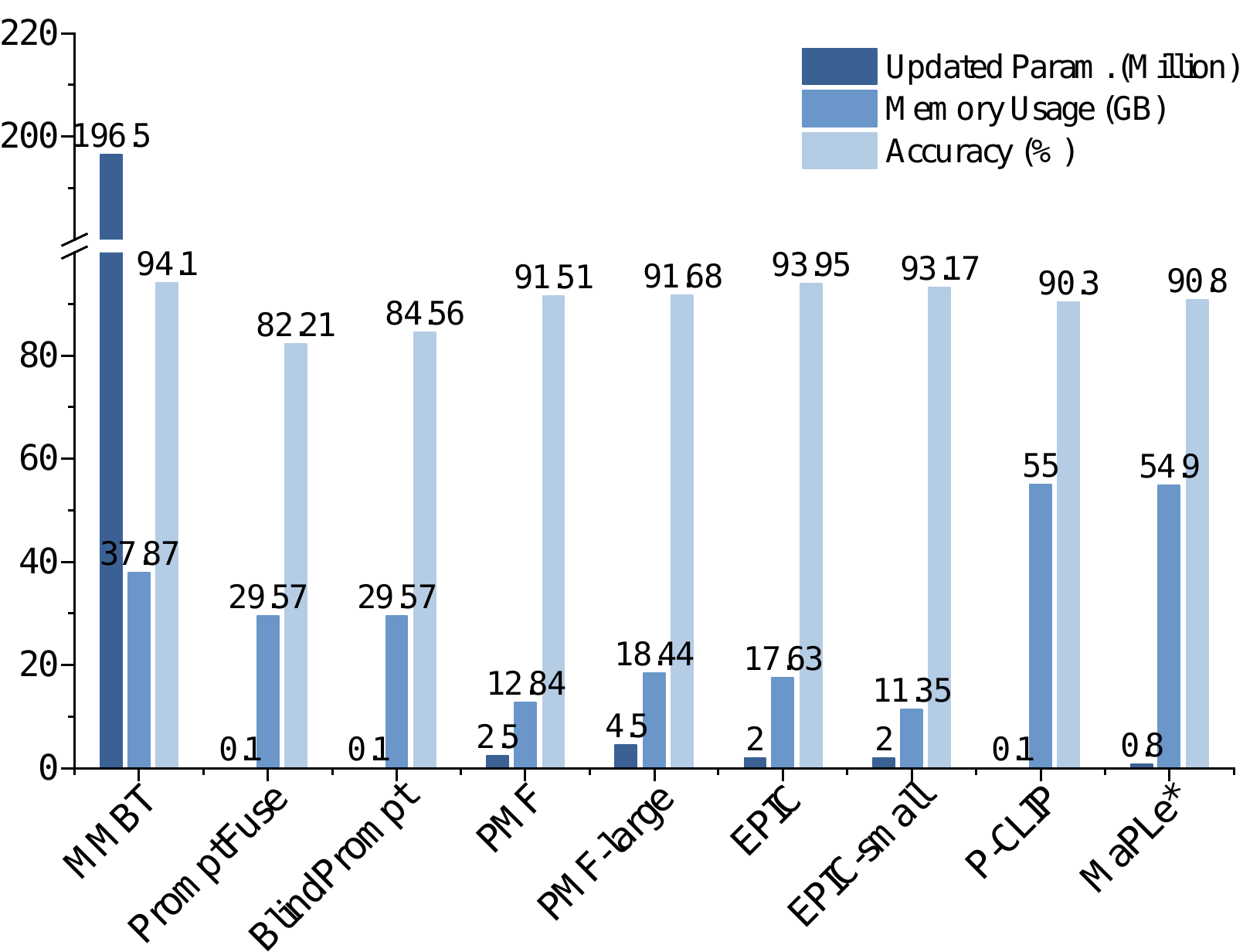}
    \caption*{(a) Comparisons on overall efficiency.}
    \end{minipage}
    \begin{minipage}{0.34\linewidth}
    \centering
    \includegraphics[width=1\textwidth]{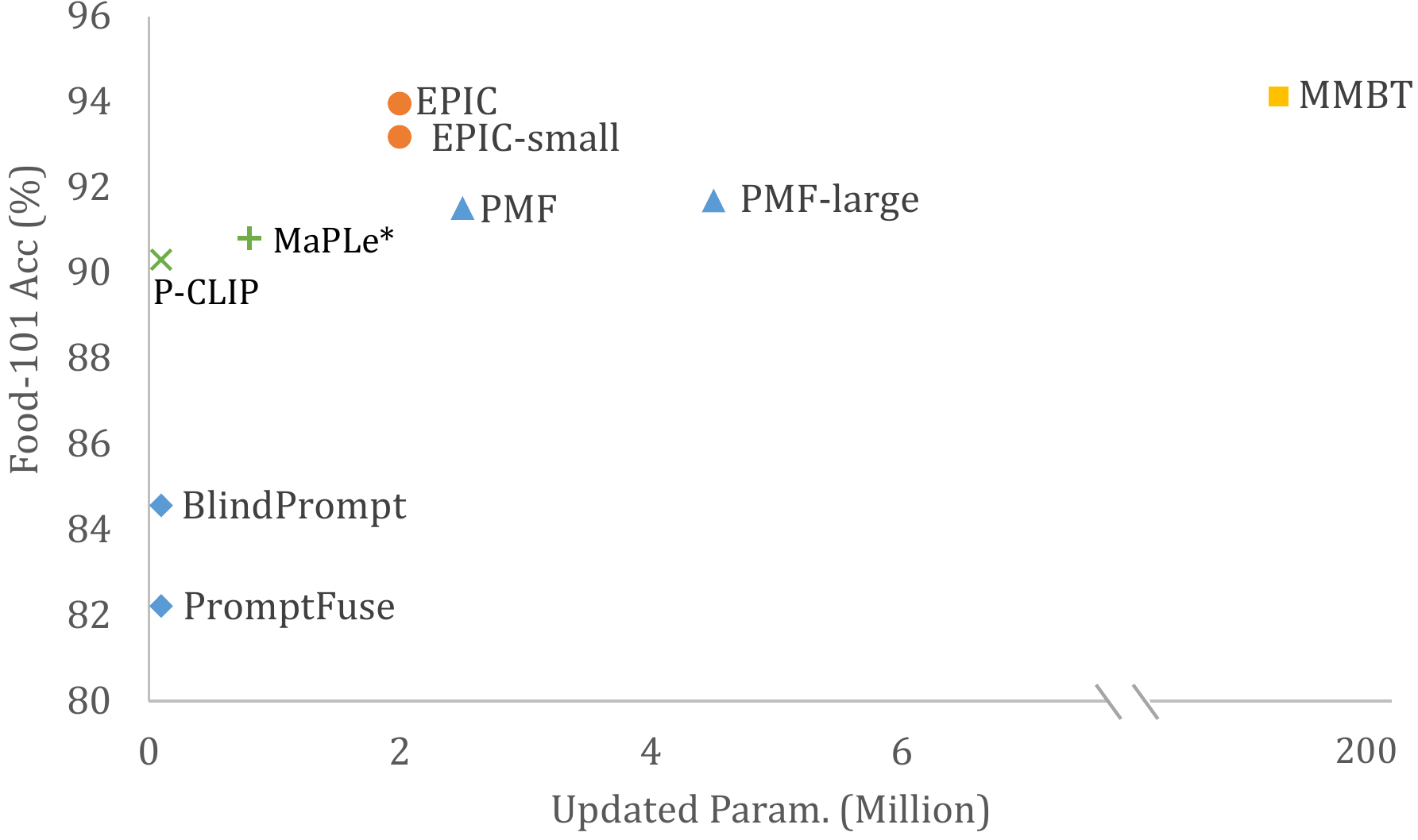}
    \caption*{(b) Comparison on efficiency of trainable parameters.}
    \end{minipage}
    \begin{minipage}{0.34\linewidth}
    \centering
    \includegraphics[width=1\textwidth]{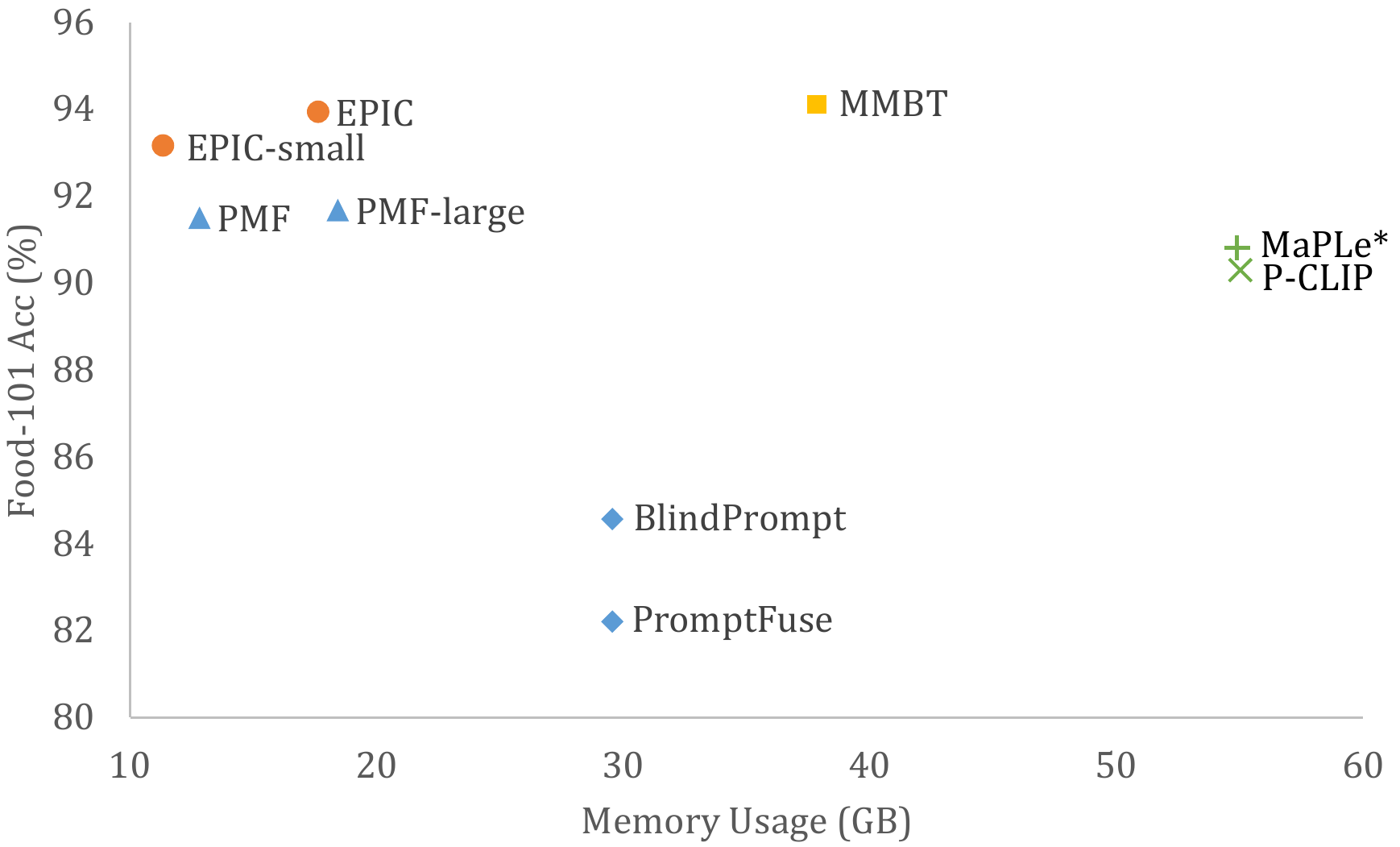}
    \caption*{(c) Comparison on efficiency of memory usage in training.}
    \end{minipage}
    \caption{The comparison of overall efficiency among EPIC and other prompt-based methods. EPIC reaches comparable performance with only 2.0M of parameters (about 1\% of the foundation model) to update and minimal GPU memory usage in training. The results are on UPMC-Food101.} 
    \label{fig5}
    \end{figure*}

	\section{Experiment}\label{sec:experiment}

	\subsection{Implementation}
    \textbf{Datasets.} We evaluate our method on three public datasets: UPMC-Food101~\cite{wang2015recipe}, MM-IMDB~\cite{arevalo2017gated}, SNLI-VE~\cite{xie2018visual}, which demonstrates efficacy and efficiency of our proposed method on three typical sub-tasks in text-image classification. \textbf{UPMC-Food101} is a popular text-image classification dataset on food,
     including 67,971 for training and 22,715 for test. \textbf{MM-IMDB} is a mainstream dataset for multi-label text-image classification on movie, containing a total of 25,956 image-text pairs. \textbf{SNLI-VE} is a visual-entailment classification dataset, in which each image-text pair includes an image premise (\(P_{image}\)) and a text hypothesis (\(H_{text}\)), with 565,286 image-text pairs in total. The task is to reason the semantic relationship (entailment, neutral, or contradiction) between \(P_{image}\) and \(H_{text}\). In this work, we only use the hypothesis description as text input, in line with prior work~\cite{li2023efficient}. 
    
    \textbf{Implementation Details.} We use pre-trained CLIP \(ViT-L/14\)~\cite{radford2021learning} as foundation model for image encoder and text encoder in frozen. Our first temporal prompt for text branch is initialized through a Gaussian distribution (\(mean=0, std=0.02\)), by which other prompts are generated. Prompt length is set to 3. We leverage 3 interaction layers empirically for the base EPIC. We also propose a small version of the proposed method, utilizing one interaction layer, namely EPIC-small.
   
    \textbf{Metrics.} We employ accuracy\( (\%) \) as metric for uni-label classification task on UPMC-Food101 and SNLI-VE datasets, and F1-micro/F1-macro for multi-label classification task on MM-IMDB dataset, following previous works~\cite{fu2022cma,li2023efficient,xue2023dynamic}.

\begin{table*}[]
    \centering
    \caption{Ablation study on components on three employed datasets. $Tmp.$ $prompt$ means temporal prompts on intermediate layers of each branch. $P. $ $interaction$ means prompt interaction through naive MLP. $Sim.$ $ strategy$ means prompt production strategies introduced in \ref{memory_hub}.}
    \begin{tabular}{l|ccc|c|c|c}
    \toprule
    &  \begin{tabular}[c]{@{}l@{}}$Tmp.$\\ prompt\end{tabular} & \begin{tabular}[c]{@{}l@{}}$P.$\\ interaction\end{tabular} & \begin{tabular}[c]{@{}l@{}} $Sim.$\\ strategy\end{tabular} & \begin{tabular}[c]{@{}l@{}} UMPC-Food101\\Acc(\%)\end{tabular} &\begin{tabular}[c]{@{}l@{}} MM-IMDB\\F1-micro/-macro\end{tabular} &\begin{tabular}[c]{@{}l@{}} SNLI-VE\\Acc(\%)\end{tabular} \\
    \midrule
    Baseline& & & & 88.80 & 60.0 / 50.8 & 68.47 \\
    P-tuning&\checkmark & & & 90.32 & 60.9 / 51.6 & 70.65\\
    Interaction&\checkmark  & \checkmark & & 92.13 & 63.5 / 54.0 & 71.89 \\
    EPIC &\checkmark & \checkmark & \checkmark & 93.95 & 65.9 / 56.3 & 73.45 \\
    
    \bottomrule
    \end{tabular}
    \label{tab:my_label4}
\end{table*}
	
	\subsection{Main Results}
    Our baseline models are as follows:\textbf{ a) }Pre-trained LMM foundation model with late fusion module, CMA-CLIP~\cite{fu2022cma}. \textbf{b)} Pre-trained LMM foundation model with prompt-tuning, MaPLe~\cite{khattak2023maple}. We re-implement MaPLe under our experimental settings on our employed image-text classification datasets. We also propose a prompt-tuning CLIP without interactions between modalities, denoted as P-CLIP.
    \textbf{c)} Existing prompt-based methods for multimodal interaction, PromptFuse~\cite{liang2022modular}, BlindPrompt~\cite{liang2022modular}, and PMF~\cite{li2023efficient} . \textbf{d)} Late fusion methods with a fusion module, \cite{narayana2019huse,liang2021multibench,hu2021unit,li1908visualbert,xue2023dynamic}.
    

    \textbf{EPIC is effective on diverse image-text classification task types.} 
    From the results, we observe that our proposed method yields superior results on uni-label classification dataset UPMC-Food101 and multimodal visual entailment dataset SNLI-VE, as well as comparable results on multi-label classification dataset MM-IMDB. In detail, we can achieve an accuracy of 93.95\(\%\) on UPMC-Food101, F1-micro/F1-macro of 65.9/56.3 on MM-IMDB, and an accuracy of 73.45\(\%\) on SNLI-VE. In particular, we outperform all existing prompt-based methods on datasets UPMC-Food101 and SNLI-VE. The small version of our proposed method (EPIC-small) also achieves competitive results on all three datasets. 
    

    \textbf{EPIC is efficient in terms of trainable parameters, and GPU memory usage.} We compare the efficiency our proposed method EPIC with other prompt-based methods, the overall comparison is presented in~\tabref{tab:my_label1} and~\figref{fig5}. 
    EPIC achieves the second highest accuracy on UPMC-Food101 with only 2.0M parameters (about 1\(\%\) of its pre-trained foundation model) to update and a minimal memory usage in training. 

\begin{figure*}
    \centering
    \begin{minipage}{0.34\linewidth} 
    \centering
    \includegraphics[width=1\textwidth]{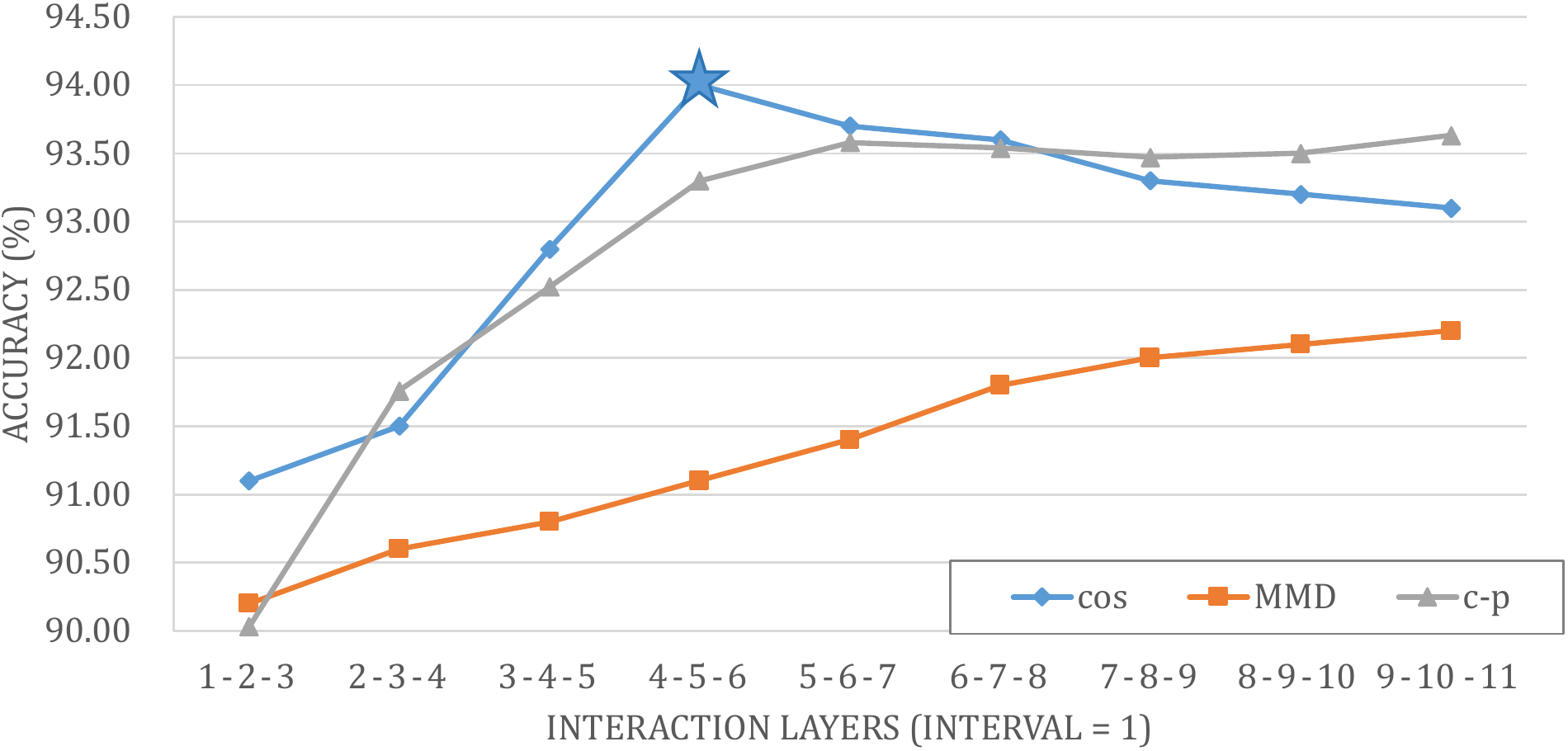}
    \caption*{(a) The results on interaction layers with interval=1.} 
    \end{minipage}
    \begin{minipage}{0.32\linewidth}
    \centering
    \includegraphics[width=1\textwidth]{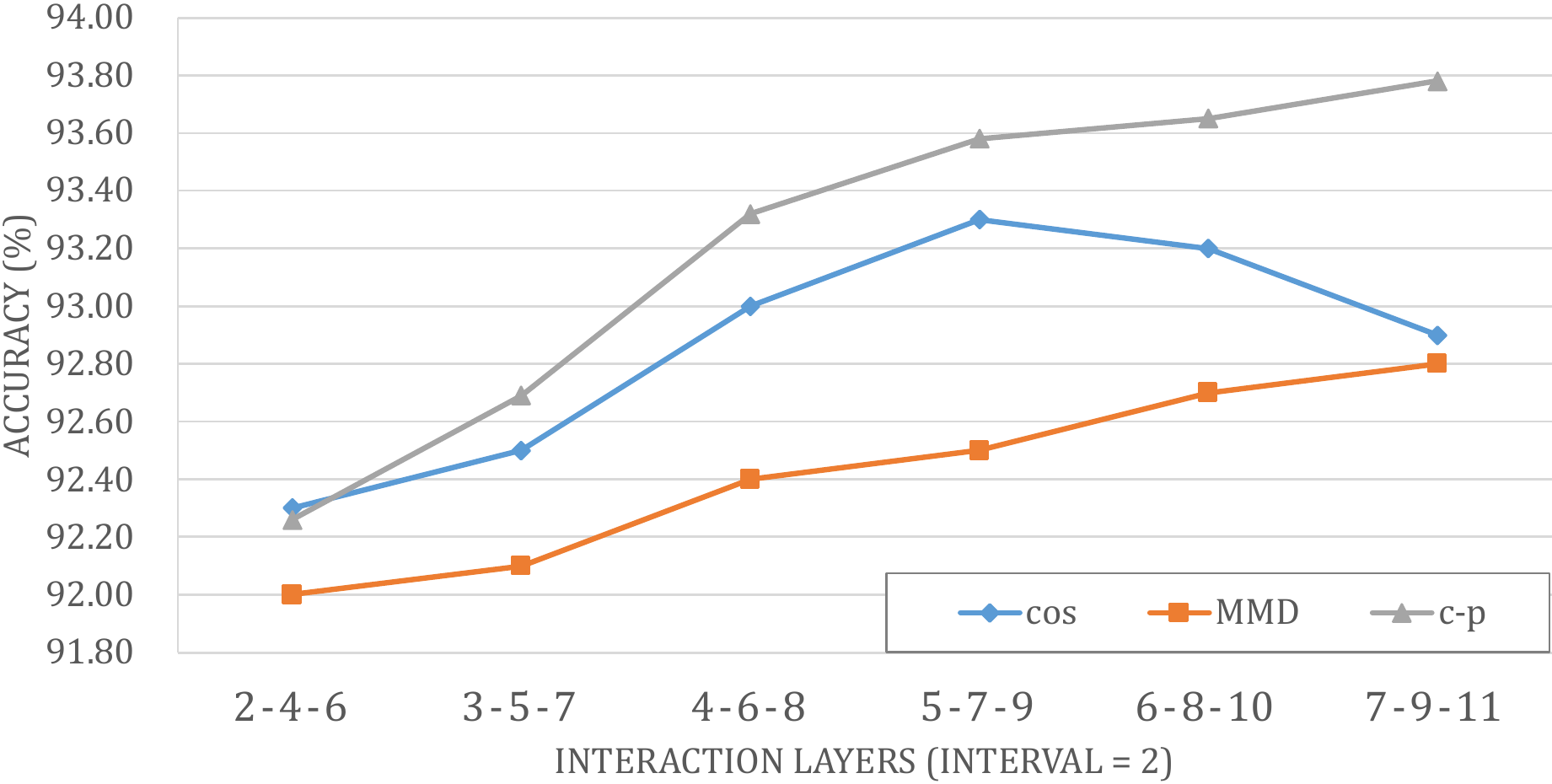}
    \caption*{(b) The results on interaction layers with interval=2.} 
    \end{minipage}
    \begin{minipage}{0.32\linewidth}
    \centering
    \includegraphics[width=1\textwidth]{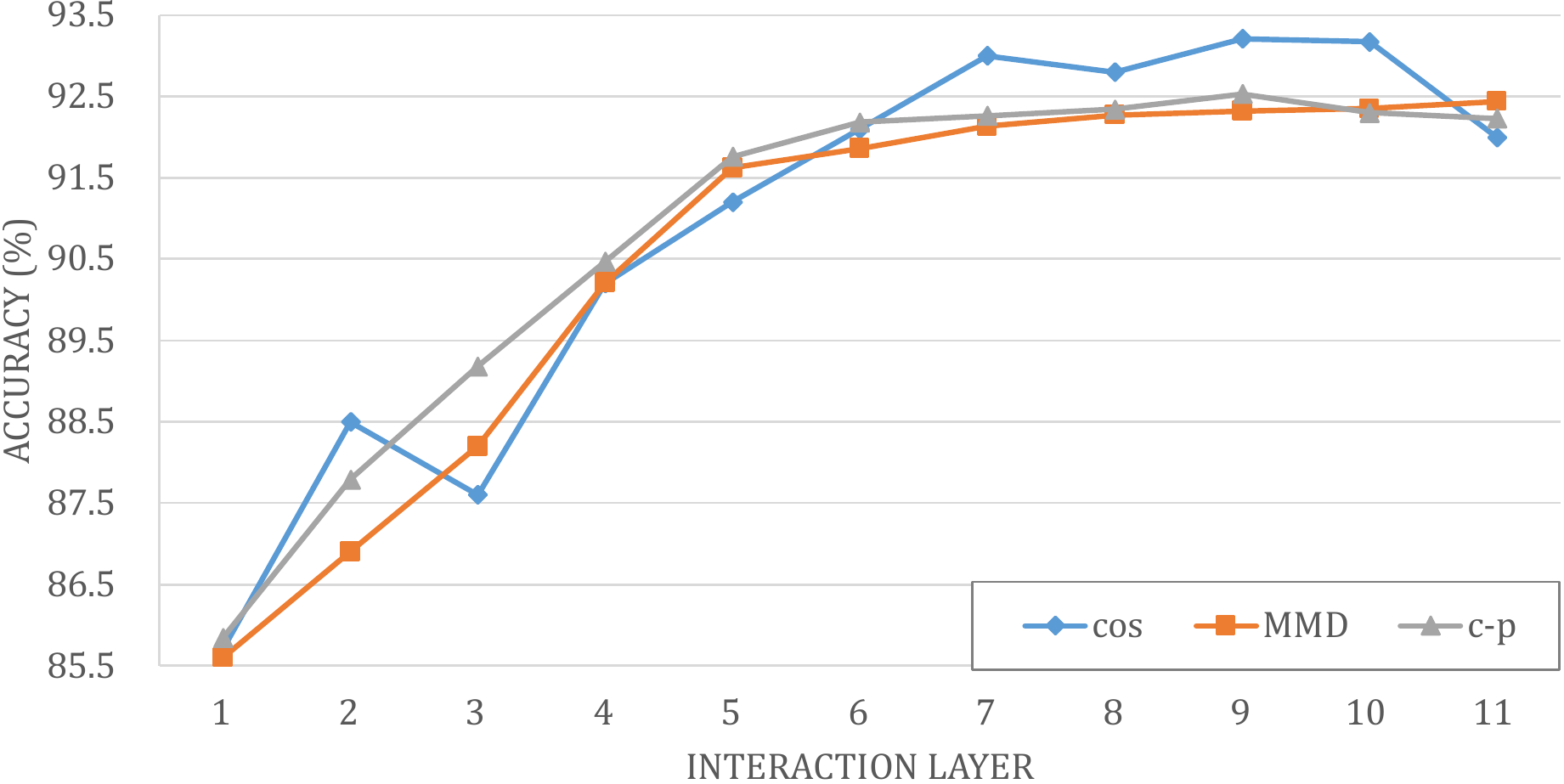}
    \caption*{(c) The results of EPIC-small on a single interaction layer.} 
    \end{minipage}
    \caption{Selection of interaction layers and similarity type. In (a) and (b), the intervals of interaction layers are selected from ${1,2}$, and similarity type is selected from $\{cos, mmd,c-p\}$, where $cos$ means cosine similarity, $mmd$ means MMD similarity, and $c-p$ denotes a combination of covariance and pearson's correlation coefficient. In (c), we perform prompt interaction on one layer. The results are on UPMC-Food101.}
    \label{fig8}
\end{figure*}

    \section{Ablation Study and Further Analysis} \label{sec:ablation}


\subsection{Ablation Study on Components}

 The baseline model does not introduce temporal prompts and is the same as the employed pre-trained foundation model. P-tuning model after leveraging temporal prompts to the baseline enhances performance. Interaction model refers to linear projection strategy of producing temporal prompts for the next layer, which concatenates temporal prompts from both modality and conducts linear projection on the concatenated vector. 
 After leveraging prompt interaction (Interaction) and similarity-based prompt production strategy (EPIC) to temporal prompts (P-tuning), the performance increases significantly, proving the effectiveness of our each component. 

\subsection{Selection of Interaction Layers and Similarity Type}
\label{interaction_layers} \label{similarity_type}
For similarity type, we utilize cosine similarity, distribution-based similarity maximum mean discrepancy (MMD), and correlation-based combination of covariance and Pearson's correlation coefficient (denoted as covariance-pearsonr), where covariance is used to measure the similarity within temporal prompt from the same modality, and Pearson's correlation coefficient is used to measure the similarity between temporal prompts from different modalities.

The experimental results are shown in~\figref{fig8}. The results indicate that the model reaches its highest performance at interaction layers (4,5,6) with cosine similarity. 
The performance of cosine similarity peaks when modality interaction is conducted at the mid-level layers, and is generally higher than that of MMD. Cosine similarity also outperforms covariance-pearsonr when interval is 1.
As for interval of interaction layers, we can see that when cosine similarity is applied, interaction layers with interval=1 generally achieve better performance; when MMD or covariance-pearsonr is applied, interaction layers with interval=2 achieve better performance. 

\section{Conclusions} \label{sec:conclusion}

In this paper, we propose EPIC, an efficient prompt interaction method for text-image classification. Our approach interacts temporal prompts on intermediate layers of the LMM foundation model.
We propose the InteractionHub module, which ensures sufficient inter-modality information exchange while maintaining low memory overhead and a manageable number of trainable parameters. We achieve the best results among prompt-based 
with the same LMM foundation model, at minimal GPU memory usage and trainable parameters. 
	
	\ifCLASSOPTIONcaptionsoff
	\newpage
	\fi

	

	\bibliographystyle{IEEEbib}
	\bibliography{main}


	
\end{document}